\documentclass[runningheads]{llncs}

\usepackage[pdftex, colorlinks=true, hyperfootnotes=true, hyperindex=true,
linkcolor=blue, citecolor=blue, urlcolor=blue]{hyperref}
\usepackage{graphicx}
\usepackage{amssymb}
\usepackage{tabularx}
\usepackage{booktabs}
\usepackage{amsmath}
\usepackage{multirow,tabularx}
\usepackage{tabularx,lipsum}
\usepackage[table]{xcolor}

\usepackage[utf8]{inputenc}
\begin{document}
\title{Evaluating Predictive Business Process Monitoring Approaches on Small Event Logs}
\titlerunning{Evaluating Predictive Process Monitoring Approaches on Small Event Logs}
%
\author{Martin Käppel\inst{1} \and Stefan Jablonski\inst{1}
\and Stefan Schönig\inst{2}\orcidID{0000-0002-7666-4482}}

\authorrunning{M. Käppel et al.}
%
\institute{Institute for Computer Science, University of Bayreuth, Germany \\
	\email{\{martin.kaeppel, stefan.jablonski\}@uni-bayreuth.de}\and
University of Regensburg, Germany\\
\email{\{stefan.schoenig\}@ur.de}\\
}
\maketitle              
\vspace{-18pt}
\begin{abstract}
Predictive business process monitoring is concerned with the prediction of how a running process instance will unfold up to its completion at runtime. Most of the proposed approaches rely on a wide number of machine learning techniques. In the last years numerous studies revealed that these methods can be successfully applied for different prediction targets. However, these techniques require a qualitatively and quantitatively sufficient dataset. Unfortunately, there are many situations in business process management where only a quantitatively insufficient dataset is available. The problem of insufficient data in the context of BPM is still neglected. Hence, none of the comparative studies investigates the performance of predictive business process monitoring techniques in environments with small data sets. In this paper an evaluation framework for comparing existing approaches with regard to their suitability for small data sets is developed and exemplarily applied to state-of-the-art approaches in next activity prediction. 

\keywords{Process Mining  \and Predictive Business Process Monitoring \and Small Sample Learning \and Process Prediction.}
\end{abstract}
\raggedbottom

\section{Introduction}\label{Sec:Introduction}
\vspace{-8pt}
Predictive business process monitoring aims at predicting of how a running process instance will unfold up to its completion at runtime based on its current state of execution.  This can help to identify problems before the process instance runs in and enables to take adequate preventive measures to avoid them. One can distinguish several prediction targets, e.g. performance predictions such as the remaining execution time \cite{RoggeSolti2015}, business rule violations \cite{Maggi2014,Metzger2015}, predictions regarding the outcome of a process instance \cite{TeinemaaBenchmark,Conforti2013}, and predictions of the next event \cite{Evermann2017} including further information as when it / they will occur and which resource(s) is / are responsible for it \cite{camargo_lstm_2019,Schonig:2018:DLP:3362381.3362411}. The majority of the proposed approaches rely on a wide number of different machine learning (ML) techniques to perform these predictions.
 
In the last years numerous comparative studies, reviews, and benchmarks of predictive business monitoring approaches have been published \cite{RamaManeiro2020,Kratsch,TeinemaaBenchmark,francescomarino2018,Verenich2019}. These studies reveal that ML techniques can be successfully applied for all the mentioned prediction tasks. However, all ML techniques are faced with the fundamental requirement of a qualitatively and quantitatively sufficient dataset. In business process management (BPM) we can have an insufficient dataset since a process \textit{(i)} is executed very seldom or has not been executed often yet, \textit{(ii)} its instances are long running, \textit{(iii)} legal regulations like the General Data Protection Regulation lead to significantly less data, or \textit{(iv)} there are fundamental changes in the intended process execution so that some historic data are not usable anymore. Especially small and medium sized companies frequently cannot fulfill this fundamental requirements of ML since not enough data is recorded. On the other hand, small data can also be a desired objective because of limited computational power or real-time feedback. Latter is, for example, characteristic for Stream Process Mining or Concept Drift. 

The problem of insufficient data in the context of BPM is still neglected \cite{Kaeppel2020SSL}. Hence, none of the comparative studies or benchmarks investigates the performance of predictive business process monitoring techniques on small data, i.e., small event logs. Hence, the  contribution of the paper is two-fold: \textit{(i)} we introduce an evaluation framework for comparing existing approaches w.r.t. their suitability for small event logs, and \textit{(ii)} analyse the suitability of existing state-of-the-art approaches in predictive business process monitoring on small event logs. This analysis is also a step towards answering the question of whether there is a lower bound for a minimum of required data for predictive business process monitoring and, if so, in which range this lower bound is located. Our results show that in many cases the algorithms allow a significant reduction of training data and, hence, training times and computational effort can be significantly reduced.

The remainder of the paper is structured as follows: In Sec. \ref{Sec:Background} we recall basic terminology and give a short introduction to the area of Small Sample Learning. Sec. \ref{Sec:Related-Work} highlights the difference between this comparative study and other surveys. In Sec. \ref{Sec:Evaluation-Framework} we describe our evaluation framework and how it can be tailored to the different areas of BPM. In Sec. \ref{Sec:Evaluation-of-Existing-Approaches-On-Small-Event-Logs} we use this framework for comparing selected state-of-the-art approaches for predicting the next activity w.r.t. their suitability for small event logs. Finally, Sec. \ref{Sec:Conclusions-and-Future-Work} outlines future work.

\vspace{-7pt}
\section{Background}\label{Sec:Background}
\subsection{Process mining}\label{SubSec:ProcessMining}
The input of process mining techniques is a \textit{(process) event log}, i.e. a set of traces of a business process (model). A \textit{trace} (also called \textit{case}) is a temporaly ordered sequence of events that are related to the same process instance. An \textit{event} is related to an activity (i.e., a step in a business process) and is characterized by various \textit{event attributes} with at least a case id ($\mathcal{C}$), the name of the corresponding activity ($\mathcal{A}$), and the timestamp of occurrence ($\mathcal{T}$). Optionally, an event contains further event attributes, such as the process participants or systems involved in executing the activity ($\mathcal{L}$) or further data payload ($\mathcal{D}_i$). Often additonal event attributes (e.g. the role of a process participant) are derived by process mining approaches and added to the events.

Let us consider the sample event log shown in Table \ref{sample-log} that provides the following event attributes: a \textit{case identifier}, the name of the executed \textit{activity}, the \textit{timestamp} of execution, the involved \textit{resource}, and two further information (\textit{amount}, \textit{key}) in form of data payload.

\begin{table}[t]
		\setlength{\belowcaptionskip}{-18pt}
	\begin{tabularx}{\textwidth}{ccXlccX}
		\toprule
		\textbf{Case ID}	& \textbf{Event ID} & \textbf{Activity}  & \textbf{Timestamp}  			& \textbf{Resource}  & \textbf{Amount}  	& \textbf{Key}  \\ \midrule
		Case1	&$e_{11}$	& A 		& 2020-10-09T14:50:17 	& MF 		&  			& SD-1 \\ 
		Case1	&$e_{12}$		& T 		& 2020-10-09T14:51:01 	& SL 		& 100 		& HG-4 \\ 
		Case1	&$e_{13}$	& W 		& 2020-11-09T12:54:39 	& KH 		&  			& HZ-2 \\ \hline
		Case2	&$e_{21}$	& A 		& 2019-04-03T08:55:38 	& MF 		&  			& SD-2 \\ 
		Case2	&$e_{22}$		& T 		& 2019-04-03T08:55:53 	& SL  		& 340 		& HK-7 \\  
		Case2	&$e_{23}$		& C 		& 2019-05-19T09:00:28 	& KH 		&  			& SGH-3 \\ \hline 
		Case3	&$e_{31}$		& A 		& 2019-11-06T10:47:35 	& MK 		&  			& SD-3  \\ 
		Case3	&$e_{32}$		& T 		& 2019-11-06T10:48:53 	& PE 		& 235 		& UG-2 \\  
		Case3	&$e_{33}$		& C 		& 2019-11-25T08:18:07 	& SJ 		&  			& KL-6 \\ \bottomrule
		
	\end{tabularx}
	\caption{Sample process event log}
	\label{sample-log}
\end{table}

\begin{definition}
	Let $\mathcal{E}$ be the set of all possible event identifiers, $\mathcal{P}$ the set of event attributes, and $\varepsilon$ the empty element. For each event attribute $p \in \mathcal{P}$, we define a function $\pi_p: \mathcal{E} \rightarrow dom\left(\mathcal{P}\right) \cup \{\varepsilon\}$ that assigns a value of the domain of $p$ to an event. However, $\varepsilon$ can only be assigned to optional event attributes.
\end{definition}
For example, for event $e_{13}$, holds $\pi_{\mathcal{A}}(e_{13}) = \texttt{"W"}, \pi_{\mathcal{C}}(e_{13}) = \texttt{"Case1"}, \pi_{\mathcal{L}}(e_{13}) = \texttt{"KH"}, \pi_{\mathcal{T}}(e_{13}) =\texttt{"2020-11-09T12:54:39"}$, $\pi_{\mathcal{D}_{\textnormal{Amount}}}(e_{13}) = \varepsilon$, and $\pi_{\mathcal{D}_{\textnormal{Key}}}(e_{13}) = \texttt{"HZ-2"}$.

\begin{definition}
Let $S$ be the universe of all traces. A \textbf{trace} $\sigma \in S$ is a finite non-empty sequence of events $\sigma = \langle e_1, ..., e_{n} \rangle$ such that for $1 \leq i < j \leq n: e_i, e_j \in \mathcal{E} \wedge \pi_{\mathcal{C}}(e_i) = \pi_{\mathcal{C}}(e_j) \wedge \pi_{\mathcal{T}}(e_i) \leq \pi_{\mathcal{T}}(e_j)$, where $|\sigma| = n$ denotes the length of $\sigma$ and $\sigma(i)$ refers to the \textit{i}-th element in $\sigma$. 
\end{definition}
This definition states that each event is unique, time within a trace is increasing, and all events with the same case identifier refer to the same process instance. 

If a process instance has finished, i.e., no additional events related to this instance are executed in the future, the trace is completed. 
\begin{definition}
	A trace $\sigma \in S$ is called \textbf{completed} if there is no $e' \in \mathcal{E}$ such that $\pi_{\mathcal{C}}(e') = \pi_{\mathcal{C}}(e)$ with $e' \notin \sigma$ and $e \in \sigma$. 
\end{definition}
An event log is defined as follows:
\begin{definition}
An \textbf{event log} $L$ is a set $L = \{\sigma_{1}, ..., \sigma_{l}\}$ of completed traces.
\end{definition}
As an example, the event log shown in Table \ref{sample-log} consists of three traces, related to the process instances Case1, Case2, and Case3. We consider traces to be equivalent with respect to one or more process perspectives (e.g. control flow) by introducing the concept of \textit{trace variants}, which defines an equivalence relation on an event log:

\begin{definition}
	Let $L$ be an event log, $\sigma_1, \sigma_2 \in L$ traces, and $\mathfrak{P} \subseteq \mathcal{P}$ a set of event attributes. We write $\sigma_1 \sim_{\mathfrak{P}} \sigma_2$, if $\sigma_1$ and $\sigma_2$ are equivalent with regard to $\mathfrak{P}$, i.e., for all $p \in \mathfrak{P}$ there is $\pi_{p}(\sigma_1(i)) = \pi_{p}(\sigma_2(i))$ for all $1 \leq i \leq max\{\vert\sigma_1 \vert, \vert\sigma_2 \vert\}$ with $\pi_{p}(\sigma(i)) = \varepsilon$ if $i > \vert \sigma \vert$. The set $\left[\sigma\right]_{\sim_{\mathfrak{P}}} := \{\sigma' \in L \vert \sigma' \sim_{\mathfrak{P}} \sigma \}$ is called a \textbf{trace variant}.  
\end{definition}
It is obvious that $\sim_{\mathfrak{P}}$ is an equivalence relation. Applying this relation to an event log provides in dependency of $\mathfrak{P}$ a more abstract or a fine grained view on the event log, since we can consider or neglect one or more process perspectives. For example applying the relation with $\mathfrak{P} = \{\mathcal{A}\}$ on the event log in Table \ref{sample-log} results in two trace variants: the trace with id Case1 represents the trace variant $\langle$A, T, W$\rangle$; a second trace variant $\langle$A, T, C$\rangle$ is represented by the two remaining traces, since they are identical with regard to the controlflow. If we additionally consider the organizational perspective, i.e. $\mathfrak{P} = \{\mathcal{A}, \mathcal{L}\}$ three trace variants evolve, since the traces with case id Case2 and Case3 differ with respect to the involved resources: $\langle$A(MF),T(SL),W(KH)$\rangle$, $\langle$A(MF), T(SL), C(KH)$\rangle$, and $\langle$A(MK),T(PE),C(SJ)$\rangle$. On the other hand, if we only consider the organizational perspective, the traces with case id Case1 and Case2 represent the same trace variant ($\langle$MF, SL, KH$\rangle$).
Usually the more event attributes are considered, the more trace variants occur since it is highly probable that they differ in one of the perspectives. We can consider the probability distribution of the trace variants within the event log.

\begin{definition}
	Let $L$ be an event log, $\Omega$ the set of trace variants with regard to $\sim_{\mathfrak{P}}$ on $L$, and $X: \Omega \rightarrow \mathbb{R}$ a discrete random variable that represents the trace variants. Then the probability for the occurence of a trace variant is given by: 
	\vspace{-7pt}
	\begin{equation*}
	P(X = \left[\sigma\right]_{\sim_{\mathfrak{P}}}) = \frac{\vert \left[\sigma\right]_{\sim_{\mathfrak{P}}} \vert}{\vert L \vert}.\vspace{-8pt}
	\end{equation*}
	We call the probability distribution of $X$ the \textbf{distribution of the trace variants}.
\end{definition}
For the example from Table 1, we obtain the following probability distribution: If only the property $\mathcal{A}$ is considered, a probability distribution of 0.33 to 0.67 follows. If additionally property $\mathcal{L}$ is regarded, three trace variants with equal probability of 0.33 result.
\vspace{-5pt}
\subsection{Small Sample Learning}\label{SubSec:Small-Sample-Learning}
\vspace{-5pt}
In recent years a new and promising area in Artificial Intelligence research called Small Sample Learning (SSL) has ermerged \cite{Shu2018}. SSL deals with ML on quantitatively inadequate data sets. This also encompass partial insufficient data sets like imbalanced datasets, where for some classes are significantly more examples than for other classes. Although, SSL has its origins in the field of computer vision, meanwhile many SSL techniques are applied in various application areas. The current SSL research is divided into two main branches: \textit{concept learning} and \textit{experience learning} \cite{Shu2018}. Experience learning attempts to solve a SSL problem by applying conventional ML techniques, by either transforming the problem into a classical ML problem through increasing the amount of data or reducing the preliminaries of the required ML algorithms. In contrast, concept learning aims to detect new concepts from only a small number of examples. Within the two branches numerous methods can be identified. Although there are numerous situations with qualitatively insufficient data in context of BPM, the application of SSL methods is still neglected in this area. This might be caused by the fact that most common SSL techniques are strongly tailored to computer vision or NLP problems and must be adapted to BPM first in order to become applicable \cite{Kaeppel2020SSL}.
\vspace{-5pt}
\section{Related Work}\label{Sec:Related-Work}
\vspace{-8pt}
This work relates to the stream of research in predictive business process monitoring and touches the area of SSL. Since, SSL methods are barely used in BPM so far, related work mainly focuses on comparative studies of existing business process monitoring approaches and the approaches itself. The problem of quantitatively insufficient data in BPM was systematically addressed the first time in \cite{Kaeppel2020SSL}, where the authors propose the idea of leveraging SSL methods for this issue. They describe their concept by the example of predictive business process monitoring, suggests SSL methods that seems promising for BPM, and describe an idea of how the effectiveness of such methods can be proven. A survey of existing SSL methods outside of BPM is presented in \cite{Shu2018}.

The problem of insufficient training data in context of predictive business process monitoring in case of next event prediction was addressed in \cite{Taymouri2020} where the authors proposed the use of Generative Adversarial Networks (GANs) for solving this issue. This network architecture outperforms other existing deep learning approaches w.r.t. accuracy and earliness of prediction. However, the authors use conventional, i.e. not small, event logs for training and evaluation. Hence, it is unclear whether it only improves results on conventional event logs or if it works for small event logs, too.

The necessity of evaluation frameworks for comparing the performance of different algorithms in BPM, respectively process mining, is not new, since the disparity of event logs, experimental setups, and different assumptions makes it often difficult to make fair comparisons \cite{RamaManeiro2020,Metzger2015}. In \cite{Rozinat2007} the authors motivate the need of an evaluation framework for process mining approaches and propose a framework for comparing model discovery algorithms. In the subfield of predictive business process monitoring there is a plenty set of different comparative studies depending on the different prediction tasks \cite{RamaManeiro2020,Kratsch,TeinemaaBenchmark,francescomarino2018,Verenich2019}. However, all of them depend on large event logs and do not consider environments with a small amount of data. Outcome-oriented techniques are reviewed and compared in \cite{TeinemaaBenchmark}, in \cite{Kratsch} with special focus on deep learning techniques. In \cite{Verenich2019} the authors give a survey and benchmark of remaining time predictions methods. In \cite{RamaManeiro2020} the authors focus on deep learning techniques for next activity prediction, activity suffix prediction, next timestamp prediction, and remaining time prediction by evaluating approaches with publicly available source code on 12 real-life event logs from various domains. 

The above discussed comparative studies observed that deep learning approaches for next activity prediction outperform classical predictions techniques, which use an explicit model representation such as Hidden Markov Models \cite{Lakshmanan}, probabilistic finite automatons or state-transition \cite{Breuker,Unuvar}. These deep learning approaches are based on different types of neural networks. Most of them use Long-Short-Term-Memory (LSTM) Neural Networks \cite{camargo_lstm_2019,Tax2017,Evermann2017,Schonig:2018:DLP:3362381.3362411}, Gated Recurrent Units (GRUs) as a variant of LSTMs \cite{Hinkka}, or Convolutional Neural Networks (CNN) \cite{Pasquadibisceglie,Mauro}. The approaches use different encoding techniques for sequences and events and consider different input data for making predictions. An overview about existing deep learning approaches including a detailed description of the underlying architectures is given in \cite{RamaManeiro2020}.

\section{Evaluation Framework}\label{Sec:Evaluation-Framework}
\vspace{-8pt}
In this section we describe the structure of our evaluation framework. We identify different challenges that must be considered by the evaluation framework.
\vspace{-8pt}
\subsection{The Issue of Small Event Logs}\label{Sec:The-issue-of-small-event-logs}
\begin{figure}[t]
			\setlength{\belowcaptionskip}{-16pt}
	\centering
	\includegraphics[scale=0.45]{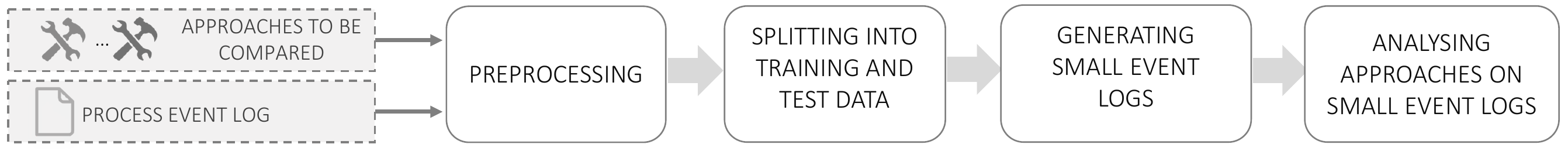}
	\caption{Conception of the evaluation framework}
	\label{abstract-concept}
\end{figure}

The evaluation framework (cf. Fig. \ref{abstract-concept}) gets two inputs: the approaches to be compared and small event logs. However, providing small event logs is a crucial challenge. This is due to the missing definition of "small event log". This question is strongly related to the question what is "big data". Also, this is still an open question in research. Due to inconsistent and sometimes contradicting definitions, that often include time dependency (i.e., define as big data what is today the largest available amount of data), domain dependency or circular reasoning (e.g. big data is the opposite to small data) this question cannot be conclusively clarified. We bypass this technical and conceptual problem by reducing event logs with various reduction factors and thereby generate small event logs of different sizes. The use of different reduction factors enables us to fully cover the broad range of "smallness", which ranges from zero or a single training example up to, for example, several 1000 examples. Hence, we are independent of concrete definitions. In a strict sense, the generated small event logs are rather "relatively small" event logs, than "small" event logs. The exact procedure of generating small event logs is described in detail in Sec. \ref{Sec:Reducing-event-logs}. 

In addition to bypassing this definition problem, generating small event logs by reducing conventional event logs has another advantage:  Since we can fall back to the non-reduced event log (so called \textit{reference log}), it is possible to compare results of an analysis achieved on a small log with the results obtained on the reference log\footnote{At this point we implicitly assume that the event logs currently used in research can be considered as quantitatively sufficient.}. This comparison allows to make quantitative statements about the impact of reducing an event log. Such a comparison is necessary to determine whether any potential loss of quality that goes in hand with the data reduction, can be better compensated by one method or by another. Hence, we measure how an approach performs depending on the reduction factor. It is likely that the achieved results also depend on the domain and structure of the considered event log.

\subsection{Preserving Comparability}\label{Sec:Preserving-comparability}
\vspace{-7pt}
When we talk about comparability in our approach, we mean to compare the prediction quality of data analysis approaches measured on reference log and the generated small event logs. For preserving comparability, it is essential to evaluate all trained models with the same test data. Hence, we first divide the event log into training and test data and afterwards we reduce only the training data and not the whole reference log. The selection of the test data is discussed in Sec. \ref{Sec:Splitting-into-training-and-testdata}. However, excluding the test data from reduction has a far-reaching consequence: Usually ML techniques, which require a split into training and test data, use training-test ratios like 80:20 or 70:30. Since we keep the test data and only reduce the training data the ratio between train and test data shifts more and more towards test data with increasing reduction factor. Hence, the less training data are used, the relatively more tests the model must pass afterwards. This seems to be unusual, however it ensures comparability and guarantees that the quality of the trained model and, as a result, the performance of the method used for training, are not overestimated but rather underestimated. In consequence, also the frequently used cross validation that enables the use of all available data for training as well for testing are not applicable anymore, since through the reduction step reference log and small event logs differ. It should be noted that dispense on comparability would avoid this shift problem but would be accompanied with interpretation problems due to meaningless splits in training and test data. Suppose that a reduced event log would only contain 10 traces left and we would split this event log into training and test data using a ratio of 80:20. Then the metrics used for evaluation would hardly be meaningful. For example the accuracy metric could only attain three different values: 0\%, 50\%, or 100\%. 

It is obvious that comparability and keeping the ratio between training and test data are diametrically opposed to each other. Since, the comparability between the reduced event logs is essential for our aims and cannot be neglected, we accept the shift of the training-test ratio in our evaluation framework. 
\vspace{-5pt}
\subsection{Reducing Event Logs}\label{Sec:Reducing-event-logs}
\vspace{-5pt}
The amount of training data is reduced by removing as many process instances from the training data such that a given reduction factor is reached. We select process instances for removal either randomly or along the time dimension. Removing instances randomly means selecting process instances randomly. When reducing along the time dimension, we order the process instances ascending by its first event timestamp and then remove the first traces according to the reduction factor. Note that the way how the process instances to be removed are selected leads to different interpretations: In case of randomly removing process instances, we simulate that a process is executed very seldom or due to legal regulations there are only a few records available for analysis. Reason for this interpretation is that the underlying time window of the event log (spanned by the earliest and latest timestamp of an event) stays nearly unchanged. However, in case of removing process instances along the time dimension, the time window is shortened. Hence, this reduction method reflects a scenario, where a process has not been executed often yet or due to fundamental changes in the intended process execution some historic data (up to a specific time) are not usable anymore. Hence, these two reduction methods are sufficient to simulate all in the introduction mentioned reasons for quantitatively insfficient event logs. Nevertheless, we implement some alternative selection methods, like removing the most recent data or removing only specific trace variants, as defined in Def. 5.

However, the reduction of the training data bears two further issues:  \textit{(i)} the possible loss of activities and resources, and \textit{(ii)} statistical bias.

\paragraph{Loss of activities and resources.}
\vspace{-8pt}
Since we generate small event logs out of large event logs, there is a risk that activities or executing units get completely lost or are finally only represented in the test data. In case of getting completely lost, the trained model would not be able to handle these activities or resources, if they occur later in productive use, since they are not encoded and therefore are unkown to the model. Therefore, we extract and buffer all occurring activities and performing units from the reference log before splitting into training and test data and before generating the small event logs. Hence, these activities and resources can be considered even if no training sample reflects them. However, this also implies that process instances in the test data that contain activities or resources that are not represented in the training data cannot be predicted well. 

\paragraph{Statistical bias.} 
\vspace{-8pt}
The reduction of the training data may be accompanied by statistical bias in the probability distribution of the trace variants (cf. Def. 6). Since most of the ML techniques are statistical methods, it affects the model quality and must therefore be considered adequately. 
Suppose that a trace variant is represented by exactly one representative in a training dataset, which consists of 100 traces. Then this trace variant has an empirical probability of 1\%. If we reduce the training data by 50\% and the representative of the considered trace variant is not removed, then the empirical probability of this trace variant increases to 2\%. At the same time other trace variants are either completely eliminated or their empirical probability decreases. In case of an even stronger reduction and under the assumption that the considered trace variant is not removed, the increase of empirical probability could be much stronger. Hence, the probability distributions of the trace variants in the considered event logs can differ significantly or reflects no longer their occurrence frequency in reality.   

However, the problem becomes less relevant the more process perspectives are considered, since process perspectives foster the singularity of traces and the number of trace variants represented in the event log tends to the number of traces in the event log. In the case that each trace variant occurs only once in the event log, removing a trace directly leads to a loss of a trace variant. This observation prohibits to reduce the event log along the probability distribution of its trace variants. However, considering this issue only from the perspective of trace variants is not sufficient. Because from the ML perspective there may be a significantly lower statistical bias since this perspective also takes the similarity between the traces into account (for example two trace variants only differ in a single event). Hence, often removing a trace variant is sufficiently compensated by another very similar trace variant that is still included in the event log. However, it is difficult to determine this compensatory effect, because it strongly depends on the considered ML technique and therefore cannot be adequately considered in the evaluation framework. 

The effects discussed above are affected by the chosen reduction method. In case of a randomly reduction, the probability distribution can be extremely distorted. The reduction along the time dimension alleviates this issue, since the probability distribution of the trace variants in a sufficiently large time window should be more similar to the distribution of the entire event log than the distribution in a randomly selected subset of the event log. This assumption also holds for the compensatory similarity effect. Hence, it is expected that via reduction along the time dimension the statistical bias can be reduced. However, it is clear, that it also depends on the particular event logs and in case of strong reduction the statistical bias cannot be longer compensated. 

\vspace{-5pt}
\subsection{Splitting into Training and Test Data}\label{Sec:Splitting-into-training-and-testdata}
\vspace{-5pt}
Still, the question remains how the event log should be split into training and test data. We use the same procedure for selecting test data as for reducing the training data, i.e., the test data is either selected randomly or along the time dimension. In the latter case, the newest process instances are used for testing and the oldest for training. The chosen split procedure may affect the achieved results, since training and test data may overlap in time by splitting the event log randomly. This could be problematic if the underlying process evolves during time and, hence, different process variants are mixed up in the event log. However, splitting along the time dimension does not prevent this issue, since the splitting can lead to a separation of a specific process variant (i.e., the test data represents a process variant that does not occur in the training data). Nevertheless, splitting along the time dimension appears closer to reality, since knowledge of the past is used to predict the future. Furthermore, the splitting along the time dimension also provides a better reproducibility.

\vspace{-5pt}	
\subsection{Architecture of the Evaluation Framework}
\vspace{-5pt}
\begin{figure}[t]
			\setlength{\belowcaptionskip}{-19pt}
	\centering
	\includegraphics[scale=0.6]{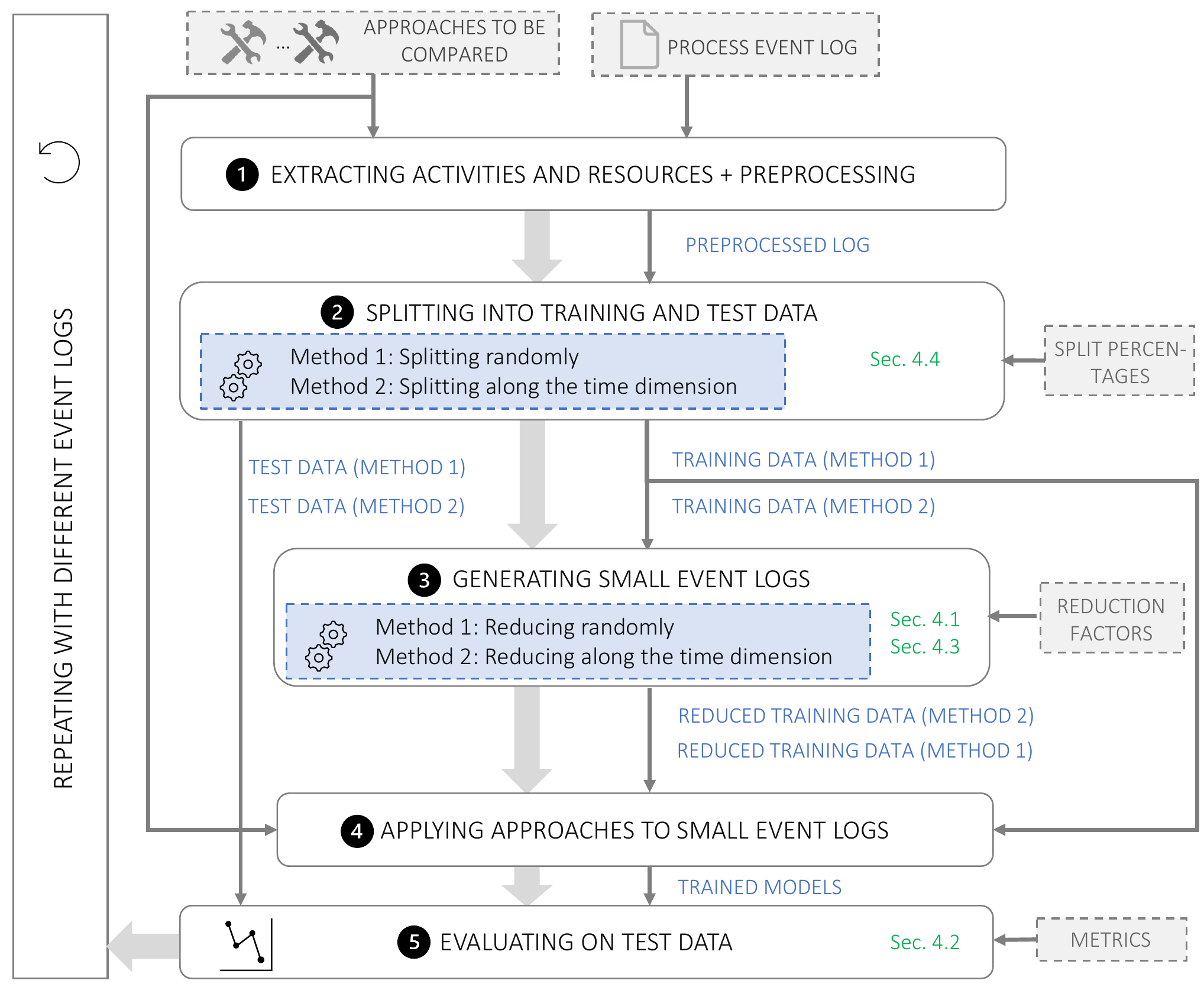}
	\caption{Architecture of the evaluation framework}
	\label{concept}
\end{figure}
In summary, the proposed evaluation framework depicted in Fig. \ref{concept} comprises five successive steps. In a first step, an event log is preprocessed depending on the selected approaches to be compared and all activities and resources of the event log are extracted and registered. Further preprocessing encompasses, for example, the removal of traces with events that have missing values. Afterwards (cf. Step 2) a preprocessed event log is split into training and test data according to one or more split ratios. The resulting training data is then reduced in the third step according to various reduction factors to generate a set of small event logs. The framework selects the traces to be removed with the same methods as used for selecting test and training data. The approaches to be compared are trained with the training data resulting from the splitting step as well as the reduced training data of the reduction step (cf. Step 4). In order to measure the performance of the considered approaches, the trained models are evaluated on the corresponding test data (cf. Step 5). For measuring the performance one or more suitable metrics are used. The selection of the metrics primarily depends on the types of approaches. After the completion of the evaluation step, all steps are repeated with further event logs to get representative results.

In general, the framework is extensible by adding alternative methods for splitting and reducing the processed data. In Fig. \ref{concept}, all parts of the framework that can be adapted and configured are marked with dashed lines. Fields with a grey background represent input parameters that have to be set to configure the framework. The blue colored fields can be extended to add more functionality to the framework. 

We conclude this section by a brief discussion whether the evaluation framework is tailored to specific process mining methods. Since the framework offers flexible preprocessing and the splitting into test and training data can be skipped it is also possible to evaluate unsupervised ML techniques. For ML techniques, which require a split into training, validation, and test data some smaller adaptions would be necessary. Since the evaluation framework does not implement any approach specific particularities, the framework can be considered as approach-agnostic.
\vspace{-12pt}
\section{Evaluation of Existing Approaches on Small Event Logs}\label{Sec:Evaluation-of-Existing-Approaches-On-Small-Event-Logs}
\vspace{-5pt}
\subsection{Dataset Description and Experimental Setup}
\vspace{-5pt}
The first three steps of the framework that are responsible for generating small event logs are implemented as a Java application. Step 4 is covered by the modified implementations of the considered approaches. Modification becomes necessary, since approaches must deal with the generated small event logs and the test data generated in the previous steps. Hence, parts in the implementations of the considered approaches that are responsible for splitting into training and test data or extracting activities and resources must be modified. Also, the approach specific evaluation components must be replaced by the evaluation component of the framework (step 5) to ensure a consistent evaluation procedure.

We evaluate our framework with a small comparative study of selected state-of-the-art approaches for next activity and role prediction. We select approaches \cite{camargo_lstm_2019} and \cite{Pasquadibisceglie} since they represent the most frequently used deep learning architectures (LSTM respectively CNN) for next activity prediction, provides publicly available source code, and achieve good results in various comparative studies. We modified the approaches in the above described way and additional changed the implementations to run with Python 3.7 and to support training on GPU. The experiments are run on a system equipped with a Windows 10 operating system, an Intel Core i9-9900K CPU3.60GHz, 64GB RAM, and a NVIDIA Quadro RTX 4000 having 6GB of memory.

We perform our experiment using 5 real-life event logs from different domains with diverse characteristics (cf. Table \ref{used-event-logs}) extracted from the \textit{4TU Center for Research Data}\footnote{https://data.4tu.nl}. For our experiment, we preprocess the event logs in the same way as it done in the considered approaches. Due to the missing resource event attribute in some traces we removed 3528 of the 13087 traces in the BPIC12 log. 
\begin{table}[t]
	\setlength{\belowcaptionskip}{-16pt}
	\begin{tabularx}{\textwidth}{p{3cm}p{1.7cm}p{1.7cm}p{1.7cm}p{1.7cm}p{1.7cm}}
		\toprule
		{\small \textbf{Event Log}}	& {\small \textbf{BPIC12}}   			& \textbf{Helpdesk}  & \textbf{BPIC13}  	& \textbf{BPIC15\_2} & \textbf{BPIC15\_5}   \\ \midrule
		Number of cases		&  9559 		   	&  4580  		& 1487  		& 832 & 1156   \\ 
		Number of activities		& 36  		  	& 14 		&  7		&   410 & 389 \\ 
		Number of roles & 8 & 4 & 4 & 11  & 10\\
		Number of events		&  	140863		&  21348		&  	6660	&   44354 & 59083\\ 
		Maximal case length & 163  & 15 & 35&  132 & 154\\
		Minimal case length & 3 & 2 & 1&   1& 5\\
		Average case length & 14.74 & 4.66 & 4.49 & 53.31  & 51.11 \\ 
		Maximal duration & 76.90 & 59.99 & 2254.85 &  1325.96 & 1343.96 \\  \bottomrule
	\end{tabularx}
	\caption{Statistic of the used event logs. Time related measures are shown in days.}
	\label{used-event-logs}
\end{table}
\vspace{-18pt}
\subsection{Tailoring the Evaluation Framework to Predictive Monitoring}
\vspace{-5pt}
We tailor the evaluation framework for evaluating predictive business process monitoring approaches in the following way. We use a training-test ratio of 70:30. The test data is selected by applying the two methods from Sec. \ref{Sec:Splitting-into-training-and-testdata}: \textit{(i)} splitting randomly, and \textit{(ii)} splitting along the time dimension. Training data are reduced according to the following reduction factors: 0.2, 0.4, 0.6, 0.8, 0.9, 0.95, and 0.99. Hereby, the traces are either removed randomly or along the time dimension. For evaluation, we use the following common metrics that can be derived from the confusion matrix\footnote{i.e. it can be calculated from true positives (TP), true negatives (TN), false positives (FP), and true positive (TP)}: \textit{(i)} \textit{Recall} (also called sensitivity) defined as $R = TP/(TP+FN)$, \textit{(ii)} \textit{Precision} $P = TP/(TP+FP)$, \textit{(iii)} \textit{F-Measure} defined as harmonic mean $F =2RP/(R+P)$ of precision and recall, and \textit{(iv)} \textit{Accuracy} defined as $A = (TP+TN)/(TP+FN+FP+FN)$. These metrics coincide with those from other comparative studies.
\vspace{-8pt}
\subsection{Results and Discussion}
\vspace{-6pt}
Due to the large number of trained models and measures and since the achieved results are comparable for all considered approaches we report in this paper only an exemplarily excerpt of the detailed measures (cf. Table \ref{eva-next-activity} and \ref{eva-next-role}) and limit ourselves to the discussion of the overall results. All further measures are provided in the repository of our implementation\footnote{https://github.com/mkaep/SSL-Evaluation-Framework}.

The most surprising observation is that only a very strong reduction factor of 95\% or 99\% significantly worsens the performance measured by different metrics. This applies for event logs where good results are achieved as well as for event logs that show poor results. A further surprising fact is that in case of a reduction along the time dimension the best results (highlighted in blue) are not achieved by the reference values (highlighted in red), i.e. often a reduced amount of training data achieves better results. However, this observation does not hold for a randomly performed reduction. This reduction method either achieves similar or slightly worse results. Hence, this observation supports the hypothesis of Sec. \ref{Sec:Reducing-event-logs} that reduction along the time dimension leads to a more representative dataset. Since, we remove those traces that have the earliest first timestamp this behaviour might also be an indicator that there is some process evolution within the logs. We also observe that the most complex event logs BPIC15\_2 and BPIC15\_5, which contain a relatively large number of activities for a comparatively small number of traces shows poor results. The reason for this behaviour seems to be the high number of activities per number of traces (cf. Table \ref{used-event-logs}). This interpretation is supported by the fact that the prediction of the next roles performs significantly better. Hence, for learning such complex logs we would need significantly more training data.

In a deeper analysis of the results we analysed whether the prediction quality between the activities differ. This analysis reveals that all trained models (also the reference models) are good in predicting frequent activities but perform poor in predicting rare activities. As a result, frequent trace variants (i.e. standard cases) are predicted very well, while rare trace variants are barely predicted correct, since the prediction model treat them as standard cases. This is somehow natural, since ML methods try to generalize the data in a simple way by neglecting rare activities. 

Hence, we can draw the following conclusions: for learning frequent trace variants of less complex logs even a significantly reduced amount of data is sufficient. For learning rare trace variants, however, it is necessary to increase the amount of data, especially by better representing rare cases. For complex logs, like the BPIC15 logs, the currently available amount of data is not sufficient, to achieve acceptable results. Hence, our results reinforce the need for SSL methods in the area of predictive business process monitoring.

\begin{table}[t]
					\setlength{\belowcaptionskip}{-5pt}
	\begin{tabularx}{\textwidth}{p{1.7cm}|X|X|X|X|X|X|X|X|X|X|X|X|X|X|X|X}
		\toprule
		& \multicolumn{8}{>{\hsize=\dimexpr8\hsize+8\tabcolsep+\arrayrulewidth\relax}c|}{\textbf{Reduced along time dimension}} & \multicolumn{8}{>{\hsize=\dimexpr8\hsize+8\tabcolsep+\arrayrulewidth\relax}c}{\textbf{Reduced randomly}}\\ \midrule
		& \multicolumn{8}{>{\hsize=\dimexpr8\hsize+8\tabcolsep+\arrayrulewidth\relax}c|}{\textit{Applied reduction factors}} & \multicolumn{8}{>{\hsize=\dimexpr8\hsize+8\tabcolsep+\arrayrulewidth\relax}c}{\textit{Applied reduction factors}}\\ \midrule
		Event Log & 0.0 & 0.2 & 0.4 & 0.6 & 0.8 & 0.9 & 0.95 & 0.99 & 0.0 & 0.2 & 0.4 & 0.6 & 0.8 & 0.9 & 0.95 & 0.99 \\ \midrule
		Helpdesk &  \textcolor{red}{0.74} & 0.74  & \textcolor{blue}{0.79}  & \textcolor{blue}{0.79}  & 0.78  & \textcolor{blue}{0.79}  & 0.78  & 0.70  & \textcolor{red}{0.74}  & 0.74  & 0.73  & 0.74  & 0.73  & 0.73  & 0.72  & 0.72 \\ 
		BPIC\_15\_5 & \textcolor{red}{0.16}  & 0.20  & 0.24  & \textcolor{blue}{0.28}  & \textcolor{blue}{0.28}  & 0.26  & 0.22  & 0.22  & \textcolor{red}{0.16}  & 0.14  & 0.13  & 0.14  & 0.09  & 0.09  & 0.08  & 0.03 \\ 
		BPIC\_15\_2 & \textcolor{red}{0.10}  & 0.09   & 0.09  & 0.08  & 0.11  & \textcolor{blue}{0.21}  & 0.17  & 0.07  & \textcolor{red}{0.11} & 0.09  & 0.09  & 0.06  & 0.08  & 0.05  & 0.06  & 0.04 \\ 
		BPIC\_13 & \textcolor{red}{0.54} & \textcolor{blue}{0.64} & 0.58 & 0.61 & 0.60 & 0.57 & 0.51 & 0.39 & \textcolor{red}{0.56} & 0.52 & 0.53 & 0.51 & 0.53 & 0.49 & 0.50 & 0.44 \\ 
		BPIC\_12 & \textcolor{red}{0.85}  & \textcolor{blue}{0.86}  & 0.85  & 0.84  & 0.85  & 0.84  & 0.84  & 0.79  & \textcolor{red}{0.85}  & 0.85  & 0.85  & 0.85  & 0.84  & 0.84  &  0.84 & 0.74  \\ \bottomrule 
	\end{tabularx}
	\caption{Accuracy measures in \% for approach \cite{camargo_lstm_2019}  (architecture "shared categorical"); \textit{Task:} next activity prediction; \textit{Split ratio:} 70:30 along time dimension.}
	\label{eva-next-activity}
\end{table}
\begin{table}[t]
				\setlength{\belowcaptionskip}{-12pt}
	\begin{tabularx}{\textwidth}{p{1.7cm}|X|X|X|X|X|X|X|X|X|X|X|X|X|X|X|X}
		\toprule
		& \multicolumn{8}{>{\hsize=\dimexpr8\hsize+8\tabcolsep+\arrayrulewidth\relax}c|}{\textbf{Reduced along time dimension}} & \multicolumn{8}{>{\hsize=\dimexpr8\hsize+8\tabcolsep+\arrayrulewidth\relax}c|}{\textbf{Reduced randomly}}\\ \midrule
		& \multicolumn{8}{>{\hsize=\dimexpr8\hsize+8\tabcolsep+\arrayrulewidth\relax}c|}{\textit{Applied reduction factors}} & \multicolumn{8}{>{\hsize=\dimexpr8\hsize+8\tabcolsep+\arrayrulewidth\relax}c|}{\textit{Applied reduction factors}}\\ \midrule
		Event Log & 0.0 & 0.2 & 0.4 & 0.6 & 0.8 & 0.9 & 0.95 & 0.99 & 0.0 & 0.2 & 0.4 & 0.6 & 0.8 & 0.9 & 0.95 & 0.99 \\ \midrule
		Helpdesk & \textcolor{red}{0.95} & \textcolor{blue}{0.95}  & 0.94  & \textcolor{blue}{0.95}  & 0.94  & \textcolor{blue}{0.95}  & 0.94  & 0.84  & \textcolor{red}{0.95}  & \textcolor{blue}{0.95}  & \textcolor{blue}{0.95}  & \textcolor{blue}{0.95}  & \textcolor{blue}{0.95}  & \textcolor{blue}{0.95} & 0.94  & 0.93 \\ 
		BPIC\_15\_5 & \textcolor{red}{0.90}  & 0.91  & 0.91  & 0.92  & \textcolor{blue}{0.93}  & 0.88  & 0.91  & 0.89  & \textcolor{red}{0.90}  & \textcolor{blue}{0.91}  & 0.90  & \textcolor{blue}{0.91}  & 0.87  & 0.87  & 0.87  & 0.52 \\ 
		BPIC\_15\_2 & \textcolor{red}{0.85}   & \textcolor{blue}{0.87}    & 0.86   & 0.86   & 0.86   & 0.82   & 0.84   & 0.55   & \textcolor{red}{0.87}  & 0.78   & 0.75   & 0.85   & 0.85   & 0.82   & 0.68   & 0.55  \\ 
		BPIC\_13 & \textcolor{red}{0.96} & \textcolor{blue}{0.96} & 0.92  & \textcolor{blue}{0.96}  & 0.95  & 0.93  & 0.83  & 0.63  & \textcolor{red}{0.97}  & 0.97  & 0.97  & 0.97  & 0.97  & 0.93  & 0.91  & 0.95  \\ 
		BPIC\_12 &  \textcolor{red}{0.97} & \textcolor{blue}{0.97}   & \textcolor{blue}{0.97}   & \textcolor{blue}{0.97}   & \textcolor{blue}{0.97}   & \textcolor{blue}{0.97}   & \textcolor{blue}{0.97}   & 0.94   & \textcolor{red}{0.97}  & \textcolor{blue}{0.97}  & \textcolor{blue}{0.97} & \textcolor{blue}{0.97}  & \textcolor{blue}{0.97} & 0.96 & 0.96 & 0.93  \\ \bottomrule 
	\end{tabularx}
	\caption{Accuracy measures in \% for approach \cite{camargo_lstm_2019}  (architecture "shared categorical"); \textit{Task:} next role prediction; \textit{Split ratio:} 70:30 along time dimension.}
	\label{eva-next-role}
\end{table}
\vspace{-13pt}
\section{Future Work}\label{Sec:Conclusions-and-Future-Work}
\vspace{-8pt}
In this paper, we propose a customizable evaluation framework for investigating predictive business process monitoring approaches w.r.t their suitability for small event logs. Our experiments reveal that training times and computational effort can be significantly reduced without any loss of quality with regard to the common metrics. For further improvement, however, it would be necessary to cope with the problem of rare trace variants. In future work the study should be extended to further approaches, event logs, and should investigate how different types of sequence and event encoding affect the performance. Furthermore, other prediction tasks, like the prediction of suffixes or the remaining time should be investigated. It is also necessary to adopt the framework for use in other subfields of process mining, like process model discovery or conformance checking.

%
%
%
%

\tiny
\bibliographystyle{splncs04}
\bibliography{literature}

\end{document}